\documentclass{article}
\usepackage[preprint]{colm2026_conference}

\usepackage[T1]{fontenc}
\usepackage{fancyvrb}
\usepackage{microtype}
\usepackage{hyperref}
\usepackage{url}
\usepackage{amsmath,amssymb}
\usepackage{booktabs}
\usepackage{graphicx}
\usepackage{subcaption}
\usepackage{wrapfig}
\usepackage{lineno}

\definecolor{darkblue}{rgb}{0, 0, 0.5}
\hypersetup{colorlinks=true, citecolor=darkblue, linkcolor=darkblue, urlcolor=darkblue}

\title{Awakening the Sleeping Agent: Lean-Specific Agentic Data Reactivates General Tool Use in Goedel Prover}

\author{Jui-Hui Chung$^{1}$, Hongzhou Lin$^{2}$, Lai Jiang$^{3}$, Shange Tang$^{1}$, Chi Jin$^{1}$ \\
$^1$Princeton Language and Intelligence, $^2$Amazon, \\
$^3$Department of Computer Science, University of Illinois at Urbana-Champaign \\
\texttt{\{juihui@princeton.edu, chij@princeton.edu\}}
}

\newcommand{\Qwen}{\textsc{Qwen3-32B}}
\newcommand{\OSS}{\textsc{GPT-OSS-120B}}
\newcommand{\Goedel}{\textsc{Goedel-Prover-V2-32B-SFT}}
\newcommand{\GoedelRL}{\textsc{Goedel-Prover-V2-32B-RL}}
\newcommand{\GoedelProver}{\textsc{Goedel-Prover-V2}}
\newcommand{\ProofNet}{\textsc{ProofNet}}
\newcommand{\MiniF}{\textsc{MiniF2F}}
\newcommand{\LeanSearch}{\textsc{LeanSearch}}

\begin{document}

\ifcolmsubmission
\linenumbers
\fi

\maketitle

\begin{abstract}
Heavy supervised fine-tuning (SFT) on a target domain can strongly suppress capabilities that were present in the base model. We study this phenomenon in formal mathematics using \GoedelProver{}, an open-source model heavily trained on 1.8M formal-math examples. After domain specialization, the model almost completely loses its ability to produce valid tool calls, even when explicitly instructed to use tools, dropping from 89.4\% function-calling accuracy in the base model to nearly 0\%. We ask whether this agentic collapse is permanent or instead reversible. To answer this question, we fine-tune the specialized model on a small amount of \emph{Lean}-specific tool-use data. Remarkably, as few as 100 agentic traces are sufficient to restore strong tool-calling behavior. Importantly, this recovery is not the result of reward hacking or benchmark-specific optimization: the recovery data is entirely drawn from the \emph{Lean} setting, where the model uses natural-language queries to search the Mathlib library for relevant theorems and lemmas, yet the regained capability transfers well beyond that domain. In particular, these same 100 \emph{Lean}-specific traces improve performance on the Berkeley Function Calling Leaderboard (BFCL) from near zero to 83.8\%, approaching the base model's 89.4\% despite the mismatch in task distribution and protocol. The recovered capability is also practically useful in-domain. On \ProofNet{}, pass@32 improves from 21.51\% to 25.81\%. Together, these results show that heavy domain SFT can suppress general tool-use ability without permanently erasing it, and that a small amount of domain-specific agentic data can awaken dormant tool-use capabilities.
\end{abstract}


\section{Introduction}
Catastrophic forgetting is a well-known risk in expert models~\citep{ kotha2024understanding, luo2025empirical}: as a general-purpose base model is heavily specialized for a target domain, capabilities outside that domain may degrade substantially. \GoedelProver{}~\citep{lin2025goedelproverv2}, a formal-math-specialized model built on top of \Qwen{} \citep{yang2025qwen3}, exhibits a notable instance of this phenomenon. Starting from the base model, \GoedelProver{} undergoes supervised fine-tuning on 1.8 million formal-math examples, followed by a reinforcement-learning stage on \emph{Lean}-only data. Despite strong in-domain performance, this specialization is accompanied by a near-complete collapse in general tool-use ability: on BFCL, accuracy falls from 89.4\% in the base model to nearly zero.

This failure raises a basic question about the nature of forgetting in specialized language models. Has heavy domain training truly erased \GoedelProver{}'s agentic capability, or has it only suppressed a capability that remains latent? Existing work has mostly approached catastrophic forgetting from a prevention perspective, showing that continual fine-tuning can impair previously acquired capabilities and proposing mitigations such as replay or pretraining-data injection during post-training \citep{kalajdzievski2024scalingforgetting,li2024revisitingcf,luo2025empirical,chaudhry2019tinyepisodic,bethune2025pretraininginjection}. At the same time, recent evidence suggests that some forgetting in LLMs may be only apparent, arising from disrupted task alignment rather than permanent capability loss \citep{zheng2025spurious}. In this work, we ask the corresponding post hoc recovery question: after agentic collapse has already occurred, can the capability be brought back? And can recovery be driven entirely by domain-specific agentic supervision, yet still generalize beyond \emph{Lean} to restore general-purpose tool use?

Our central finding is that surprisingly little domain-specific agentic data is sufficient to induce broad generalization. With only 100 \emph{Lean}-specific agentic traces, the model improves substantially not only on tool-augmented \ProofNet{} theorem proving, but also on general-purpose function calling on BFCL. This cross-domain transfer is the key result: although the supervision is entirely confined to the \emph{Lean} setting, its effect extends well beyond formal mathematics, improving general tool calling capability. 

This efficiency is practically important. High-quality agentic traces are significantly more expensive to construct than standard single-turn SFT data, as they require tool orchestration, execution, response parsing, and filtering. The fact that just 100 domain-specific traces can drive broad gains suggests that even highly specialized models retain a latent capacity for general tool use, which can be unlocked through targeted supervision.

\paragraph{Contributions.}
\begin{list}{$\bullet$}{%
  \setlength{\leftmargin}{1.5em}%
  \setlength{\itemindent}{0pt}%
  \setlength{\labelsep}{0.5em}%
  \setlength{\labelwidth}{1em}%
}
\item We identify \emph{agentic collapse} in \GoedelProver{}, showing that heavy specialization for formal mathematics nearly eliminates general-purpose tool-use behavior: on BFCL, function-calling accuracy drops from 89.4\% in the base model to nearly zero.

\item We address a key data bottleneck: general-purpose models can call tools but are weak in \emph{Lean}, while specialized formal-math models are strong in \emph{Lean} but weak at tool use. We resolve this mismatch with a \emph{cross-model trace distillation} pipeline that combines general-purpose agentic behavior with domain-specific generation quality to construct high-quality \emph{Lean}-specific traces (Section~\ref{sec:method}).

\item We show that a very small amount of \emph{domain-specific} agentic data can induce broad and unexpected cross-domain generalization. As few as 100  \emph{Lean}-specific agentic traces improve not only \ProofNet{} pass@32 from 21.51\% to 25.81\%, but also BFCL non-live accuracy from 4.6\% to 83.8\%, even though the training data deos not contain any general-purpose APIs or JSON-style function calls (Sections~\ref{sec:recovery} and~\ref{sec:bfcl}).

\item We provide both quantitative and qualitative analysis of the resulting behavior (Section~\ref{sec:new_theorems}), showing that improved retrieval grounding enables the use of theorems that the model cannot generate independently.
\end{list}

\section{Can Model Merging Bring Back Tool Use?}
\label{sec:model_merging}

We study the problem in a post hoc setting: the model has already been heavily specialized for a narrow domain, and the question is whether general-purpose tool use can still be brought back afterward. This question arises when post-training preserves strong domain competence while suppressing behaviors needed for another domain, such as agentic interaction. \GoedelProver{} is a concrete example of this tradeoff: it is strong at \emph{Lean} theorem proving, yet nearly loses general-purpose tool use. In this setting, model merging is a natural first baseline. Since the base model, \Qwen{}, retains strong function-calling ability while \GoedelProver{} retains strong \emph{Lean} proving ability, a natural question is whether interpolating between the two checkpoints can recover both capabilities at once.

\paragraph{Setup.}
We test whether model merging can recover \emph{in-domain} tool-use behavior, where the model is explicitly instructed to call tools while solving \emph{Lean} theorem-proving problems. Concretely, we evaluate SLERP interpolation~\citep{10.1145/325334.325242,yang2026model} between \Goedel{} and \Qwen{}. We measure tool use by asking, for each problem, whether at least one of the $k$ sampled generations produces a valid executed tool call; we report the resulting problem-level \emph{tool-call compliance}. We sweep SLERP merge weights from 0\% base (pure \Goedel{}) to 100\% base (pure \Qwen{}) in 10\% increments, and evaluate each merged checkpoint on \MiniF{}~\citep{zheng2021minif2f} and \ProofNet{}~\citep{azerbayev2023proofnet}.

\begin{figure}[t]
  \centering
  \begin{subfigure}[t]{0.48\linewidth}
    \centering
    \includegraphics[width=\linewidth]{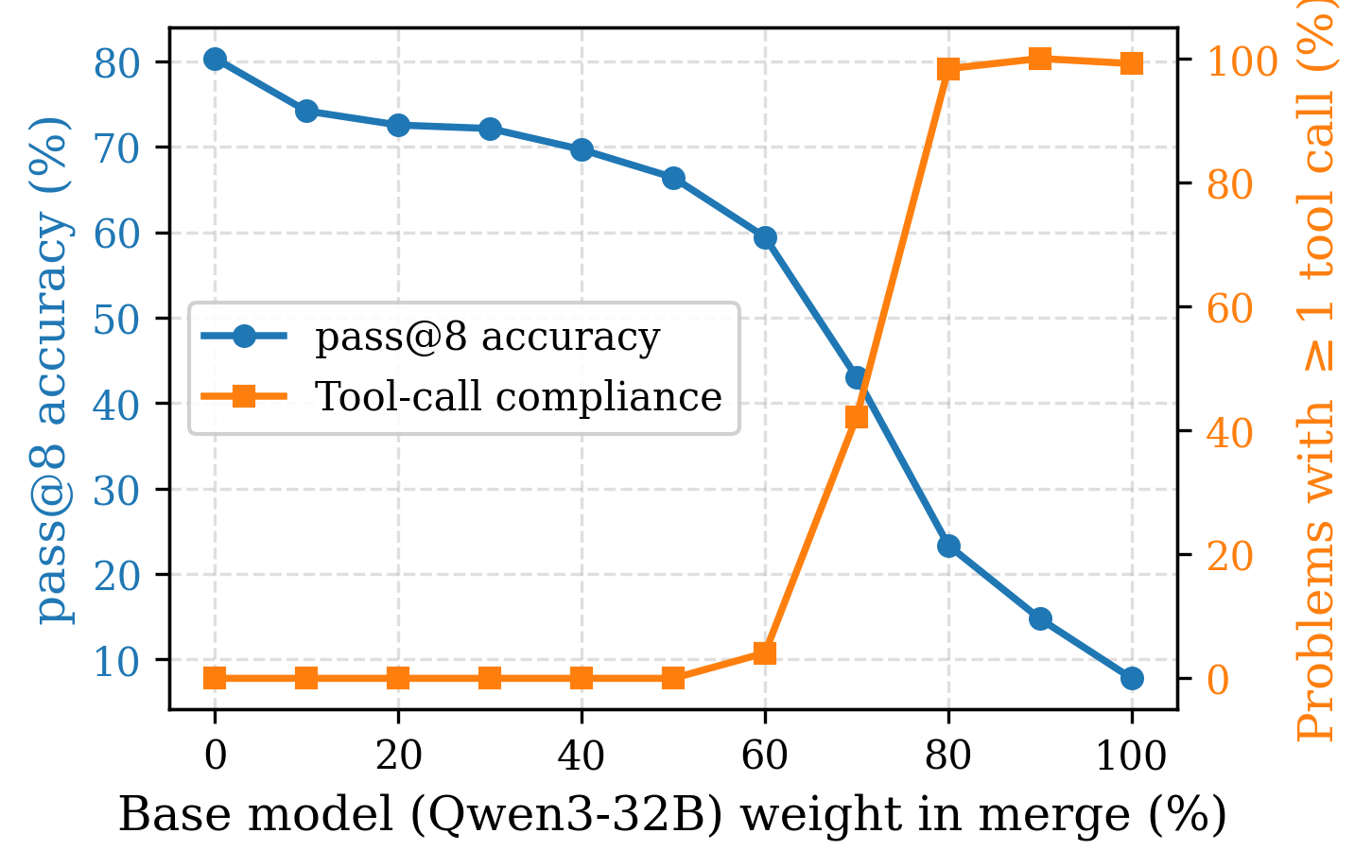}
    \caption{\MiniF{} (244 problems, pass@8). }
    \label{fig:minif2f_slerp}
  \end{subfigure}
  \begin{subfigure}[t]{0.48\linewidth}
    \centering
    \includegraphics[width=\linewidth]{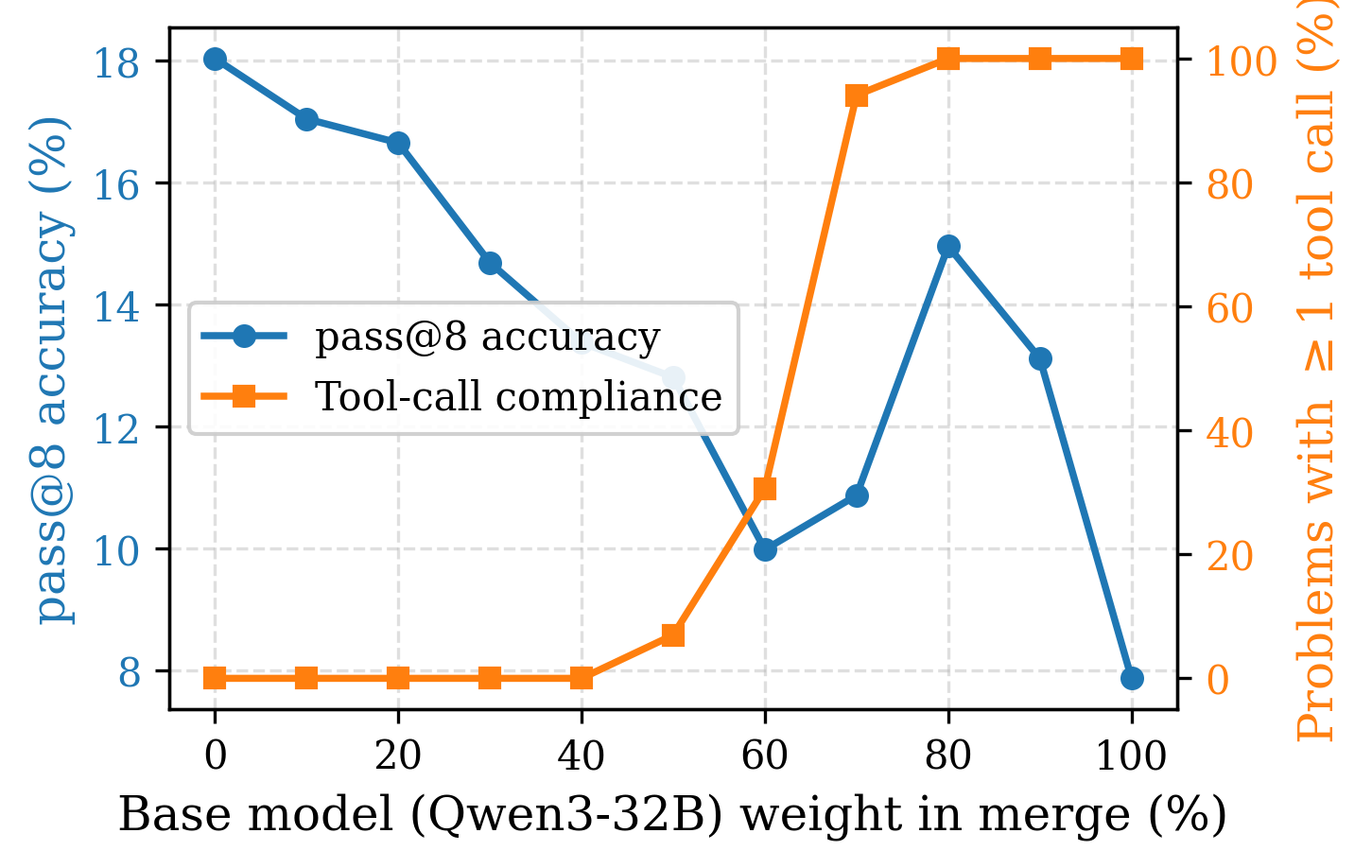}
    \caption{\ProofNet{} (186 problems, pass@8). }
    \label{fig:mention_vs_toolcall}
  \end{subfigure}
  \hfill

    \caption{SLERP merge sweeps on two Lean theorem-proving benchmarks. Left axis (blue): evaluation performance on the Lean task (on \MiniF{} and \ProofNet{}). Right axis (orange): problem-level tool-call compliance. Across both benchmarks, increasing the base-model weight consistently improves tool-calling behavior while degrading in-domain Lean performance. The overall tradeoff is very clear: as tool-use capability returns, evaluation accuracy on the Lean task drops substantially. This shows that naive merging does not jointly preserve strong theorem-proving ability and strong tool-calling behavior.}

  \label{fig:slerp_sweeps}
  \vspace{-0.3cm}
\end{figure}

\paragraph{Discussion and connection to prior work.}
The main takeaway from Figure~\ref{fig:slerp_sweeps} is that naive model merging does not preserve strong \emph{Lean} proving and strong tool use at the same time. As the merge shifts toward the base model, tool-calling behavior reappears, but in-domain theorem-proving performance declines. Although prior work has shown that model merging can mitigate forgetting in some transfer settings~\citep{alexandrov2024mitigating}, our results indicate that this strategy is insufficient to maintain both behaviors simultaneously. We therefore connect this phenomenon within the broader literature on catastrophic forgetting and continual learning.

\vspace{-0.3cm}
\paragraph{Catastrophic forgetting and continual learning.}
Catastrophic forgetting refers to the degradation of previously learned behaviors after training on new data \citep{mccloskey1989catastrophic,french1999catastrophic}. In the continual-learning literature, it has been studied extensively in sequential task settings, where common mitigation strategies include replay-based methods, regularization, and parameter-isolation approaches \citep{chaudhry2019tinyepisodic,kirkpatrick2017ewc,shi2025continualsurvey}. These methods are primarily designed to \emph{prevent} forgetting as new tasks are introduced. Recent work shows that large language models exhibit similar forgetting during post-training and continual fine-tuning \citep{luo2025empirical, wang2023trace, kalajdzievski2024scalingforgetting,li2024revisitingcf,zheng2025spurious}. These findings suggest that forgetting in LLM post-training is not merely a classical continual-learning artifact, but a practical challenge for maintaining broad capabilities.

More recent work has begun to ask not only how forgetting can be mitigated, but also whether some apparent forgetting reflects true loss at all. On the mitigation side, approaches include replay with synthetic or curated data \citep{huang-etal-2024-mitigating}, as well as injecting a small fraction of pretraining data during fine-tuning to stabilize general capabilities \citep{bethune2025pretraininginjection}. Closely related, the notion of \emph{spurious forgetting} suggests that some post-training capability loss may not correspond to permanent erasure of the underlying competence, but instead to a disruption in task alignment that makes the capability behaviorally inaccessible \citep{zheng2025spurious}. This framing is particularly relevant in our setting: although the specialized model appears to almost completely lose general-purpose tool use, our later results indicate that this behavior can be brought back.

Our setting differs from prior work in two key ways. First, we study near-complete collapse rather than mild or gradual degradation: after specialization, \GoedelProver{} exhibits almost no tool-call compliance. Second, our focus is explicitly \emph{post hoc} and centered on \emph{generalization}: rather than asking how to prevent forgetting during training, we ask what can still be done after specialization is complete, and whether narrowly domain-specific supervision can reinstate behavior that transfers beyond formal mathematics. Closest to our setting, REAL-Prover likewise builds on supervised fine-tuning for \emph{Lean} theorem proving, but it focuses on constructing a retrieval-augmented prover rather than on the post hoc question we study here: whether suppressed tool-use behavior can be reinstated after specialization~\citep{shen2025real}. This emphasis also differs from recent systems such as Hilbert and Numina-Lean-Agent, which improve formal reasoning by incorporating stronger external reasoning components or by directly building on frontier general-purpose coding agents~\citep{varambally2025hilbert,liu2026numina}.

\section{Cross-Model Data Curation for Lean agentic trajectories}
\label{sec:method}

A central challenge in our setting is that no available model can generate high-quality agentic traces end-to-end. The behaviors we need---accurate Lean theorem proving and correct tool use---are split across different models. General-purpose models such as \Qwen{} and \OSS{},  can plan tool use, emit valid tool calls, and incorporate retrieved results, but they perform poorly on Lean code generation. In contrast, the specialized model \GoedelProver{} produces strong Lean proofs, but has largely lost the ability to participate in tool-augmented interaction loops.

As a result, neither model alone can produce the kind of training data required for our setting: trajectories that interleave tool calls with correct Lean code. This creates a data bottleneck. To address it, we construct agentic training traces by combining the complementary strengths of the two models, using a cross-model distillation pipeline described below.

\begin{figure}[h!]
  \centering
  \includegraphics[width=0.95\linewidth]{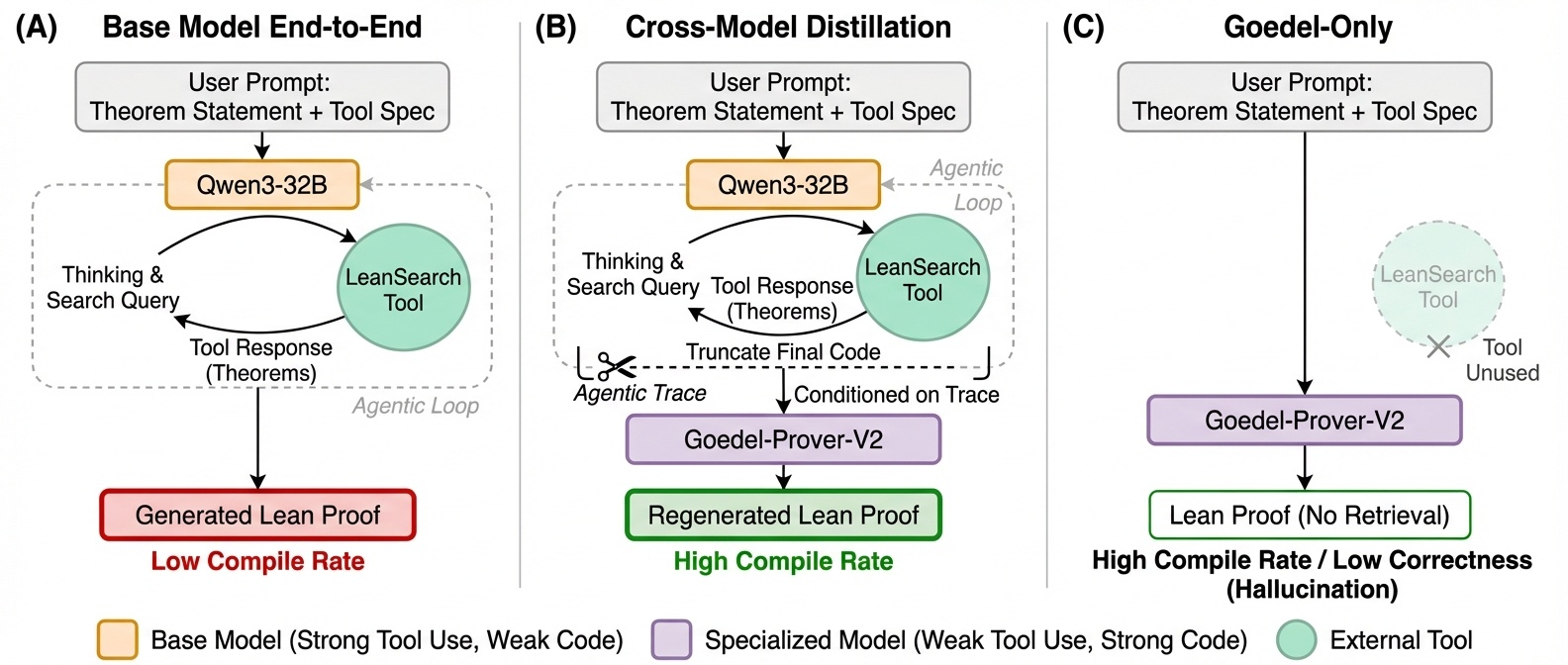}
  \caption{Cross-model trace distillation pipeline. \textbf{(A)}~\Qwen{} runs the full agentic loop (tool calls + Lean code) but produces low-quality code. \textbf{(B)}~We keep \Qwen{}'s agentic prefix and have \Goedel{} regenerate the final proof, then filter for compilation success and retrieval grounding. \textbf{(C)}~\Goedel{} alone generates strong Lean code but skips tool use entirely.}
  \label{fig:pseudo_pipeline}
\end{figure}

\paragraph{Cross-model data curation for Lean agentic traces.}
To address the data bottleneck, we construct training data by combining the complementary strengths of the two models (Figure~\ref{fig:pseudo_pipeline}). Starting from roughly 250k Lean-formalized Open Math Reasoning (OMR) problems~\citep{moshkov2025aimo}, our pipeline consists of four stages:

\begin{list}{\arabic{enumi}.}{%
  \usecounter{enumi}%
  \setlength{\leftmargin}{1.5em}%
  \setlength{\itemindent}{0pt}%
  \setlength{\labelsep}{0.5em}%
  \setlength{\labelwidth}{1em}%
}
\item \textbf{Agentic trace generation.} 
We use \Qwen{} to generate tool-augmented interaction traces over Lean problems, exposing a \LeanSearch{} tool and prompting the model to call it whenever additional retrieval appears helpful.

\item \textbf{Prefix extraction.} 
From each interaction, we retain the tool-use portion of the trajectory while discarding the final proof produced by \Qwen{}, since its Lean generations are typically low quality and compile only about 0.4\% of the time.

\item \textbf{Proof regeneration.} 
We next prompt \GoedelProver{} with the retained agentic context to generate the final Lean proof. By combining retrieved context with \GoedelProver{}'s stronger Lean ability, this step produces substantially more successful Lean proofs, raising the compilation rate from 0.46\% to 21\%.

\item \textbf{Quality filtering.} 
Finally, we retain only trajectories whose regenerated proof compiles successfully and that explicitly use at least one theorem returned by the \LeanSearch{} tool, yielding a curated set of high-quality Lean agentic examples for training.
\end{list}

This pipeline yields 18K high-quality agentic traces. The two models contribute complementary strengths: \Qwen{} provides the agentic scaffolding and retrieval behavior, while \GoedelProver{} provides high-quality Lean code generation.

\section{Experimental Results}
\label{sec:results}

\subsection{Experimental Setup: SFT on top of \GoedelProver{}}
\label{sec:setup}

\paragraph{Training data.}
From the 18K curated Lean agentic traces described in Section~\ref{sec:method}, we construct training sets of size 100, 1,000, and the full 18K examples.
The smaller subsets are not random samples: we first filter the full pool to remove traces that contain \texttt{sorry}/\texttt{admit}, lack any \LeanSearch{} calls, or consist only of trivial tactic lookups (e.g., queries for \texttt{simp}, \texttt{linarith}).
From the filtered pool, we apply quality-weighted sampling that favors traces where more retrieved theorems are actually used in the final proof, queries are diverse across retrieval rounds, and tool interaction is deeper (multiple calls).
The sample is stratified by the number of tool calls (1, 2, 3, 4+) to ensure representation of both simple and multi-step retrieval patterns, and capped per Mathlib namespace to encourage topical diversity.
We then conduct SFT with these data on top of the domain specialized model \Goedel{} and \GoedelRL{}. This allows us to measure how much domain-specific agentic data is needed to induce broad gains.

\vspace{-0.1cm}
\paragraph{Models.}
We consider two checkpoints: \Goedel{}, the SFT checkpoint, and the final RL-trained checkpoint \GoedelRL{}{}. Both are domain-specialized continuations of \Qwen{} \citep{yang2025qwen3}. \Goedel{} is trained on 1.79 million formal-math SFT examples, and the RL checkpoint is further optimized on Lean-only data. In our main experiments, we apply SFT with the constructed agentic data on top of both checkpoints. This lets us test whether behavior differs from different stages of specialization.

\vspace{-0.1cm}
\paragraph{\LeanSearch{} tool.}
\LeanSearch{} is a retrieval tool over Lean's mathlib that returns relevant theorems given a natural-language or formal query \citep{gao2024semantic}. We index mathlib declarations with informal descriptions in a ChromaDB vector database using \texttt{intfloat/e5-mistral-7b-instruct} embeddings. At inference time, each \LeanSearch{} query is embedded with a retrieval-specific instruction and matched against the index, returning the top-$k$ results with theorem names, formal and informal statements.

\vspace{-0.1cm}
\paragraph{Agentic data formatting.}
Each training example is a multi-turn dialog following an interleaved reasoning--action--observation loop similar to ReAct \citep{yao2022react}. A system message specifies the \LeanSearch{} tool schema and invocation format, a user message provides the Lean theorem statement, and the assistant may perform one or more rounds of reasoning followed by tool calls. Tool responses are returned as user messages, and the final assistant message contains the Lean proof.
\vspace{-0.1cm}
\paragraph{Training hyperparameters.}
All agentic SFT runs use full-parameter SFT with LLaMA-Factory \citep{zheng2024llamafactory} on 32 NVIDIA H100 80\,GB GPUs (4 nodes $\times$ 8 GPUs) with DeepSpeed ZeRO-3 and CPU offloading.
We use a per-device batch size of 1 with 8 gradient accumulation steps, yielding an effective batch size of 256.
The learning rate is $1 \times 10^{-5}$ with a cosine schedule and 10\% warmup ratio.
We set the maximum sequence length to 16{,}384 tokens and train in bf16 precision.
The number of epochs is scaled inversely with dataset size to keep total training tokens roughly comparable: 128 epochs for 100 examples, 32 epochs for 1K examples, and 1 epoch for the full 18K set.
These hyperparameters are shared across all experiments on both the SFT checkpoint (\Goedel{}) and the RL checkpoint (\GoedelRL{}).

\subsection{Agentic SFT Improves Lean Proving Benchmarks: \MiniF{} and \ProofNet{}}
\label{sec:recovery}
\label{sec:analysis}

Our setup is strictly post-hoc: starting from already specialized checkpoints, we apply another stage of supervised fine-tuning (SFT) with Lean agentic data, without revisiting the original domain-specialization data. Concretely, we continue training from existing \Goedel{} and \GoedelRL{} checkpoints using Lean agentic datasets of size 100, 1K, and 18K, and then evaluate the resulting models on the in-distribution Lean theorem-proving benchmarks \MiniF{} and \ProofNet{}.

\begin{table}[t]
  \centering
  \caption{\ProofNet{} (186 problems) and \MiniF{} (244 problems) results in a strictly post-hoc setting. Starting from already specialized checkpoints (\Goedel{} or \GoedelRL{}), we continue supervised fine-tuning with Lean agentic data of size 100, 1K, or 18K, without rerunning the original specialization pipeline. We report standard pass@$k$ performance together with retrieval-grounding statistics. ``Retr.'' denotes the proportion of compiled proofs that used at least one theorem returned by \LeanSearch{}.}
  \label{tab:proofnet_main}
  \small
  \begin{tabular}{@{}lrrrrrr@{}}
    \toprule
    & \multicolumn{3}{c}{\ProofNet{} (186 problems)} & \multicolumn{3}{c}{\MiniF{} (244 problems)} \\
    \cmidrule(lr){2-4} \cmidrule(lr){5-7}
    Model variant & pass@8 & pass@32 & Retr. & pass@8 & pass@32 & Retr. \\
    \midrule
    \Qwen{} & 8.60 & 13.44 & 96.0 & 7.79 & 14.34 & 31.4 \\
    \midrule
    \multicolumn{7}{l}{\textit{Panel A: SFT checkpoint (\Goedel{})}} \\
    \midrule
    \Goedel{} & 17.74 & 21.51 & 0.0 & 80.33 & 84.02 & 0.0 \\
    \midrule
    + 100 Lean agentic SFT & 20.97 & 25.81 & 93.8 & 77.46 & 81.97 & 78.5 \\
    + 1K Lean agentic SFT & 23.66 & 27.96 & 92.3 & 79.92 & 82.79 & 83.2 \\
    + 18K Lean agentic SFT & 24.73 & 27.96 & 88.5 & 79.92 & 85.25 & 74.5 \\
    \midrule
    \multicolumn{7}{l}{\textit{Panel B: Post-RL checkpoint (\GoedelRL{})}} \\
    \midrule
    \GoedelRL{} & 16.67 & 22.58 & 0.0 & 84.02 & 87.70 & 0.0 \\
    \midrule
    + 100 Lean agentic SFT & 24.73 & 26.88 & 90.0 & 77.87 & 83.20 & 80.3 \\
    + 1K Lean agentic SFT & 23.12 & 26.88 & 94.0 & 81.15 & 84.43 & 84.5 \\
    + 18K Lean agentic SFT & 21.51 & 26.88 & 90.0 & 79.92 & 84.02 & 72.2 \\
    \bottomrule
  \end{tabular}

\end{table}

Table~\ref{tab:proofnet_main} shows that post-hoc agentic SFT consistently improves \ProofNet{}, while producing only mild changes on \MiniF{}. Notably, even a very small amount of agentic data already leads to a large improvement. With SFT on only 100 agentic examples, pass@32 on \ProofNet{} rises from 21.51\% to 25.81\%, while scaling the agentic dataset to 1K and 18K continues to improve performance, reaching 27.96\%. As we show later in Section~\ref{sec:new_theorems}, this gain is driven by the model leveraging the search tool to access more challenging theorems that it cannot generate on its own, helping explain the larger improvement on \ProofNet{}.

Beyond the aggregate gains, post-hoc agentic SFT also produces a clear behavioral shift. In the post-hoc agentic SFT models, tool-call execution becomes near-universal: 99.1\%--99.9\% of generations contain an executed \LeanSearch{} call. The retrieved theorems are then carried into the final proofs. For example, in the 100-example run, 45 out of 48 pass@32-solved \ProofNet{} problems (93.8\%) contain at least one retrieved theorem identifier in the compiled Lean proof. This shows that the model grounds its Lean reasoning in externally retrieved lemmas rather than relying solely on internal recall.

A related pattern appears in Panel~B for the post-RL checkpoints. Here, post-hoc agentic SFT with all three Lean agentic data sizes (100, 1K, and 18K) reaches the same pass@32 on \ProofNet{}, corresponding to 50 solved problems. However, the solved sets are not identical across runs: 40 problems are shared, while the remaining successes are drawn from a smaller pool of harder problems with low per-generation success rates. One plausible hypothesis is that RL induces overspecialization by concentrating the model’s proof-generation policy on a narrower subset of training-preferred behaviors. This can reduce coverage over the more diverse proof trajectories required by \ProofNet{}, which may help explain why the post-RL checkpoint starts slightly higher than the SFT checkpoint before recovery (22.58\% vs.\ 21.51\%) but saturates at a lower post-hoc level (26.88\% vs.\ 27.96\%).

Overall, these results show that a small amount of domain-specific agentic SFT is sufficient to induce a robust behavioral shift, enabling the model to integrate retrieval into formal reasoning while preserving its underlying domain expertise.

\subsection{Retrieval Provides Out-of-Model Information}
\label{sec:new_theorems}

The previous section shows that retrieved theorems are frequently incorporated into compiled proofs. A natural question is whether retrieval simply \emph{surfaces} information already accessible to the model, or whether it supplies \emph{out-of-model} information that the model cannot produce on its own. To make this distinction precise, we define the model's \emph{generative support}: the set of all theorem identifiers that the base \Goedel{} model ever emits in its outputs, aggregated over all \ProofNet{} problems and all samples (including both compiled and non-compiled generations). This provides an empirical characterization of what the model can express from its parameters under standard decoding.

We then classify each retrieved-and-used theorem in a compiled RAG proof as follows:
(i) \emph{in-model} if its identifier appears in the generative support, and
(ii) \emph{out-of-model} otherwise.
By construction, out-of-model theorems are never produced by the model in any baseline run without access to the Search tool, and therefore represent information that is not accessible through the model's internal generation alone.

Table~\ref{tab:genuinely_new} presents the results. Across all three post-hoc agentic SFT models, roughly half of retrieved-and-used theorem occurrences are out-of-model. For the +100 model, 45.4\% of retrieved-and-used occurrences (63 unique theorems) are never produced by the base model, and 36 of 48 solved problems contain at least one such theorem. This proportion remains stable across data scales (45--50\%), indicating a consistent structural effect rather than a byproduct of increased data.

\begin{table}[t]
  \centering
\caption{Analysis of retrieved theorem usage on \ProofNet{} at pass@32 after agentic SFT on \Goedel{} SFT checkpoint. A retrieved theorem is \emph{in-model} if its identifier appears in the generative support of the base \Goedel{} model, without any tool calling, and \emph{out-of-model} otherwise. ``Retr.\ \& used'' counts retrieved theorem occurrences referenced in compiled proofs. ``Out-of-model'' and ``In-model'' report the counts in each category, with unique theorem counts in parentheses. ``Solved w/ out'' is the number of pass@32-solved problems whose compiled proof contains at least one out-of-model retrieved theorem.}
  \label{tab:genuinely_new}
  \small
\begin{tabular}{@{}lccccc@{}}
  \toprule
  \multicolumn{6}{l}{\textit{Panel A: \Goedel{} SFT checkpoint + Lean agentic SFT}} \\
  \midrule
Variant & Used retr. & Out-of-model & In-model & \% Out-of-model & Solved w/ out \\
  
  \midrule
  +100  & 790 & 359 (63 uniq.) & 431 (86 uniq.) & 45.4 & 36/48 \\
  +1K   & 739 & 373 (59 uniq.) & 366 (76 uniq.) & 50.5 & 35/52 \\
  +18K  & 745 & 358 (58 uniq.) & 387 (80 uniq.) & 48.1 & 36/52 \\
  \bottomrule
\end{tabular}
\end{table}

The two categories of retrieved theorems play qualitatively different roles. In-model theorems are mostly basic results that the model already generates reliably on its own, such as \texttt{mul\_assoc}, \texttt{le\_trans}, \texttt{one\_mul}, and \texttt{smul\_eq\_zero}. In these cases, retrieval appears to reinforce or stabilize capabilities the model already possesses. By contrast, out-of-model theorems are more specialized Mathlib lemmas that the model never generates in baseline runs without access to the Search tool. Examples from the 100-example post-hoc agentic SFT model include:

\begin{list}{$\bullet$}{%
  \setlength{\leftmargin}{1.5em}%
  \setlength{\itemindent}{0pt}%
  \setlength{\labelsep}{0.5em}%
  \setlength{\labelwidth}{1em}%
}
\item \texttt{FiniteField.prod\_univ\_units\_id\_eq\_neg\_one}: Wilson's theorem.
\item \texttt{solvable\_quotient\_of\_solvable}: solvability of quotient groups.
\item \texttt{IsNilpotent.isUnit\_one\_add}: unit construction from nilpotent elements.
\item \texttt{Subgroup.normal\_inf\_normal}: normality of intersections.
\item \texttt{IsLocalMin.deriv\_eq\_zero}: Fermat's theorem on local extrema.
\item \texttt{Ideal.mul\_eq\_inf\_of\_coprime}: coprime ideal factorization.
\end{list}

These examples illustrate where retrieval provides a genuine capability gain. The model formulates a natural-language query, uses the Search tool to surface the correct lemma name and signature, and then incorporates that lemma into a proof that would otherwise fail. This shows that the gains in Table~\ref{tab:proofnet_main} are not merely due to format-level behavioral recovery, but instead arise from access to mathematical knowledge beyond the model's internal generation.




\begin{table}[t]
  \centering
  \caption{BFCL v4 function-calling accuracy (\%). ``Simple'' is the mean over simple function-calling tasks in Python, Java, and JavaScript. ``Non-Live'' and ``Live'' are  means over their respective four functional categories, excluding irrelevance, and ``Overall'' averages these two. Panel~A starts from the SFT checkpoint \Goedel{}, while Panel~B starts from the post-RL checkpoint \GoedelRL{}. All post-hoc agentic SFT models are trained only on Lean-specific \LeanSearch{} traces (Section~\ref{sec:method}), with no exposure to general-purpose APIs or BFCL-style function schemas.}
  \label{tab:bfcl}
  \label{tab:bfcl_rl}
  \small
  \begin{tabular}{@{}lrrrrrrrr@{}}
    \toprule
    Model & Simple & Mult. & Par. & P.M. & Non-Live & Live & Overall & Irrelev. \\
    \midrule
    \Qwen{} & 77.25 & 96.00 & 93.00 & 91.50 & 89.44 & 81.87 & 85.66 & 80.00 \\
    \midrule
    \multicolumn{9}{l}{\textit{Panel A: SFT checkpoint (\Goedel{})}} \\
    \midrule
    \textsc{Goedel-Prover-V2-SFT} & 4.25 & 9.00 & 3.50 & 1.50 & 4.56 & 6.14 & 5.35 & 95.83 \\
    \midrule
    + 100 agentic SFT & 74.83 & 92.50 & 86.50 & 81.50 & 83.83 & 72.91 & 78.37 & 21.67 \\
    + 1k agentic SFT & 75.00 & 94.00 & 85.50 & 79.50 & 83.50 & 72.91 & 78.21 & 36.25 \\
    + 18k agentic SFT & 69.33 & 88.50 & 82.50 & 80.50 & 80.21 & 70.32 & 75.27 & 49.17 \\
    \midrule
    \multicolumn{9}{l}{\textit{Panel B: Post-RL checkpoint (\GoedelRL{})}} \\
    \midrule
    \textsc{Goedel-Prover-V2-RL} & 30.50 & 25.00 & 13.50 & 3.50 & 18.13 & 18.73 & 18.43 & 91.25 \\
    \midrule
    + 100 agentic SFT & 77.25 & 94.00 & 88.00 & 84.00 & 85.81 & 69.20 & 77.51 & 21.67 \\
    + 1k agentic SFT & 73.83 & 92.00 & 89.50 & 87.50 & 85.71 & 74.91 & 80.31 & 33.75 \\
    + 18k agentic SFT & 71.50 & 94.00 & 88.50 & 81.50 & 83.88 & 73.64 & 78.76 & 34.58 \\
    \bottomrule
  \end{tabular}
\end{table}

\subsection{Generalization Beyond: Lean Agentic Data Reactivates General Tool Use Capability}
\label{sec:bfcl}

We next evaluate whether Lean-specific agentic SFT transfers beyond the training domain using the Berkeley Function Calling Leaderboard (BFCL v4; \citealt{patil2025bfcl}), which measures structured tool use across diverse, non-mathematical APIs such as weather, databases, and file systems. This evaluation is fully out-of-distribution: both the tools and output formats (JSON) differ from the Lean-specific XML-style calls used during training. Importantly, all SFT data used in our paper is strictly Lean-specific, consisting only of \LeanSearch{}-based traces, with no exposure to general-purpose APIs or function-calling schemas.

As a first observation, we find that \textbf{agentic collapse extends beyond Lean}. Before agentic SFT, the specialized model fails almost completely on BFCL. In particular, \Goedel{} achieves near-zero function-calling accuracy, showing that the loss of tool-use behavior is not confined to Lean-specific settings.

\paragraph{Lean-specific supervision transfers to general tool use.}
Table~\ref{tab:bfcl} shows surprisingly strong out-of-distribution transfer: despite being trained only on Lean-specific agentic traces, the model recovers strong function-calling performance on BFCL. Remarkably, as few as 100 Lean agentic examples are sufficient to produce a large gain. This recovery is especially striking because the training and evaluation settings differ completely in both tools and format: the model is trained on Lean-specific XML-style tool calls, yet generalizes successfully to JSON-based APIs across diverse domains. Taken together, these results show that domain-specific agentic supervision does not merely teach a particular tool schema, but can reactivate a more general capability for structured tool use.

\paragraph{Rapid recovery with minimal data.}
Another striking pattern is how little data is needed to recover strong BFCL performance. Using only 100 Lean agentic examples already restores most of the function-calling ability, while increasing the dataset to 1K or 18K provides only limited additional benefit and can even slightly reduce performance. This pattern is inconsistent with standard scaling behavior and suggests that the model is not learning a new capability from scratch. Instead, it supports a suppression-based view: the underlying tool-use capability remains present in the base model, but domain specialization suppresses its expression, and a small amount of agentic supervision is sufficient to unlock it again.

\paragraph{Irrelevance--compliance trade-off.}
We also observe a clear trade-off between function-calling recovery and irrelevance accuracy. As tool-call compliance improves, irrelevance detection drops from 95.83\% for \Goedel{} to 21.67\% (+100), 36.25\% (+1K), 49.17\% (+18K), and 67.08\% (+120K). A likely explanation is overfitting in the action policy: our Lean agentic training data contains only positive tool-use examples and does not include irrelevant or no-tool cases. As a result, the model is biased toward calling a tool whenever one is available, even in cases where abstention would be the correct behavior. This pattern suggests that recovering irrelevance handling likely requires targeted training with explicit no-tool or irrelevant-tool examples, rather than more positive tool-use supervision alone.

\section{Limitations}

Our BFCL results suggest that broad agentic behavior is only partially reinstated. While single-turn function calling improves substantially and approaches the base model overall, the model still exhibits large errors on irrelevance detection. In particular, the base model achieves 80.00\% accuracy on irrelevance, whereas post hoc agentic SFT models remain far below this level (e.g., 21.67\% with 100 traces and 49.17\% with 18k traces in Panel~A). This indicates that the model has not fully recovered the ability to withhold tool calls when no tool is needed, and instead tends to over-call tools even when inappropriate.

In addition, we do not yet fully understand the differing behavior between SFT and RL checkpoints. Although both can be improved via agentic SFT, they exhibit distinct recovery trajectories. This suggests that different post-training procedures may suppress agentic behavior in qualitatively different ways, an issue that remains underexplored. 


\section{Conclusion}
Domain-specialized models that cannot use tools leave significant capability gains on the table.
We show that heavy SFT on 1.79M formal-math examples does not permanently destroy agentic capabilities, even when it completely suppresses them.
As few as 100 agentic traces recover tool-call compliance from 0\%, yield +4.30 points on \ProofNet{} pass@32, and restore general function-calling on BFCL to 94\% of base-model capability despite containing no BFCL-relevant data.
The practical recipe is simple: heavy domain SFT with abundant cheap data, then minimal agentic reactivation with expensive but tiny data.

Our evidence indicates that domain SFT \emph{suppresses} rather than \emph{destroys} agentic capabilities.
If tool-calling knowledge were truly erased, we would expect recovery to scale linearly with data, no cross-protocol transfer, and no residual tool awareness---yet all three predictions are contradicted.
Recovery saturates at ${\sim}$100 examples; Lean-specific XML reactivation recovers JSON-based function-calling on BFCL (Section~\ref{sec:bfcl}); and the SLERP sweep reveals that models \emph{mention} tools long before they can \emph{execute} them (Section~\ref{sec:model_merging}).
These findings suggest that tool calling and domain expertise are \emph{modular} capabilities with distinct saturation points, consistent with the LIMA hypothesis \citep{zhou2023lima} extended to the tool-use setting.
More broadly, our cross-model distillation pipeline---using a general-purpose model for agentic scaffolding and a specialized model for domain generation---applies wherever these complementary strengths exist, and the BFCL results suggest that any setting requiring structured tool interactions is vulnerable to the same collapse under heavy SFT.

\section*{Ethics Statement}
This work studies capability recovery in large language models specialized for formal mathematics. The models, datasets, and tools used are all publicly available or derived from publicly available resources. Our research does not involve human subjects, private data, or applications with direct potential for societal harm. The primary societal implication of our findings is that heavy domain specialization may suppress but not erase latent capabilities in LLMs; we believe this observation is important for the community to understand both the robustness and the fragility of post-training behavior. All experiments were conducted on 32 NVIDIA H100 GPUs; we acknowledge the associated energy consumption but note that the fine-tuning runs are small relative to pretraining-scale compute.

\section*{Reproducibility Statement}
We take several steps to support reproducibility. All base and specialized models used in this work are publicly available: \Qwen{}~\citep{yang2025qwen3} and \GoedelProver{}~\citep{lin2025goedelproverv2}. Full training hyperparameters---including learning rate, batch size, number of epochs, sequence length, and optimizer configuration---are reported in Section~\ref{sec:setup}. The fine-tuning pipeline uses the open-source LLaMA-Factory framework~\citep{zheng2024llamafactory} with DeepSpeed ZeRO-3. Our retrieval tool is built on publicly available components (\texttt{intfloat/e5-mistral-7b-instruct} embeddings and ChromaDB). Evaluation on BFCL follows the official benchmark protocol. For \ProofNet{} and \MiniF{}, we report pass@$k$ metrics with the sampling parameters specified in Section~\ref{sec:setup}. We will release the agentic trace dataset and fine-tuning code upon publication.

\bibliography{refs}
\bibliographystyle{colm2026_conference}

\appendix
\section{Tool-calling instruction prompt}
\label{app:tool_prompt}

The following system message is prepended to every agentic training example and used at inference time to specify the \LeanSearch{} tool schema and invocation format.

\begin{Verbatim}[frame=single, fontsize=\scriptsize, framesep=2mm]
You are a Lean 4 theorem prover that uses leansearch tool to find relevant theorems
in Mathlib before producing the Lean 4 code.

# Tools

You may call one or more functions to assist with the user query.

You are provided with function signatures within <tools></tools> XML tags:
<tools>
{"type": "function", "function": {"name": "leansearch", "description": "Search
Mathlib (Lean4) for theorems relevant to a natural language math query. Return a
concise list of theorem names with signatures.", "parameters": {"type": "object",
"properties": {"query": {"type": "string", "description": "Informal math query
describing the statement to search for."}}, "required": ["query"]}}}
</tools>

For each function call, return a json object with function name and arguments within
<tool_call></tool_call> XML tags:
<tool_call>
{"name": <function-name>, "arguments": <args-json-object>}
</tool_call>
\end{Verbatim}

\section{Representative agentic trace}
\label{app:trace_example}

Below is an abbreviated agentic training example for IMO 1959 Problem~1, illustrating the multi-turn format described in Section~\ref{sec:method}. The system message is shown in Appendix~\ref{app:tool_prompt}. We abbreviate chain-of-thought blocks with \texttt{[...]} to highlight the structural elements: reasoning, tool calls, tool responses, and the final Lean proof.

\begin{Verbatim}[frame=single, fontsize=\scriptsize, framesep=2mm]
<|im_start|>system
[System prompt from Appendix A]<|im_end|>

<|im_start|>user
Complete the following Lean 4 code:

```lean4
import Mathlib
import Aesop
set_option maxHeartbeats 0
open BigOperators Real Nat Topology Rat

/-- Prove that the fraction 21n+4 / 14n+3 is irreducible for every natural number n.-/
theorem imo_1959_p1 (n : N) (h0 : 0 < n) :
    Nat.gcd (21 * n + 4) (14 * n + 3) = 1 := by sorry
```

Before producing the Lean 4 code to formally prove the given theorem, provide a
detailed proof plan outlining the main proof steps and strategies. The plan should
use the leansearch tool to find relevant theorems from Mathlib that could help
prove this theorem formally.<|im_end|>

<|im_start|>assistant
<think>
I need to prove gcd(21n+4, 14n+3) = 1. Applying the Euclidean algorithm:

  21n + 4 = 1*(14n + 3) + (7n + 1)
  14n + 3 = 2*(7n + 1) + 1

So gcd = 1. Equivalently, the linear combination
  -2*(21n+4) + 3*(14n+3) = 1
proves coprimality. In Lean's Nat (no negatives), rearrange as:
  3*(14n+3) = 2*(21n+4) + 1

[... detailed exploration of Bezout coefficients and Lean formalization strategies ...]

Let me search Mathlib for a helper lemma that converts a linear combination into
a gcd=1 proof.
</think>

<tool_call>
{"name": "leansearch", "arguments": {"query": "Prove that gcd(21n+4, 14n+3) = 1
for all natural numbers n using the Euclidean algorithm or linear combinations."}}
</tool_call><|im_end|>

<|im_start|>user
<tool_response>
{"tool": "leansearch", "top_k": 10, "theorems": [
  "Tactic.NormNum.nat_gcd_helper_2' (x y a b : N)
      (h : x * a = y * b + 1) : Nat.gcd x y = 1",
  "Tactic.NormNum.nat_gcd_helper_1' (x y a b : N)
      (h : y * b = x * a + 1) : Nat.gcd x y = 1",
  "Int.gcd_dvd_iff {a b : Z} {n : N} :
      gcd a b | n <-> exists x y : Z, n = a * x + b * y",
  ...
]}
</tool_response><|im_end|>

<|im_start|>assistant
<think>
The retrieved theorem nat_gcd_helper_1' states: if y*b = x*a + 1 then gcd(x,y) = 1.
Setting x = 21n+4, y = 14n+3, a = 2, b = 3 gives exactly
  (14n+3)*3 = (21n+4)*2 + 1
which holds since both sides equal 42n+9. This lets us conclude
gcd(21n+4, 14n+3) = 1 directly, without needing gcd commutativity.

[... verification of argument ordering and theorem applicability ...]
</think>

```lean4
import Mathlib
import Aesop
set_option maxHeartbeats 0
open BigOperators Real Nat Topology Rat

/-- Prove that the fraction 21n+4 / 14n+3 is irreducible for every natural number n.-/
theorem imo_1959_p1 (n : N) (h0 : 0 < n) :
    Nat.gcd (21 * n + 4) (14 * n + 3) = 1 := by
  have h_eq : (14 * n + 3) * 3 = (21 * n + 4) * 2 + 1 := by ring
  exact Tactic.NormNum.nat_gcd_helper_1' (21 * n + 4) (14 * n + 3) 2 3 h_eq
```
\end{Verbatim}

\section{Case studies: out-of-model theorem retrieval}
\label{app:out_of_model_traces}

We present two abbreviated inference traces from the +100 reactivation model on \ProofNet{}, illustrating cases where the model retrieves and uses a Mathlib theorem that lies entirely outside its generative support (Section~\ref{sec:new_theorems}). In both traces, the retrieved theorem is never produced by the base \Goedel{} model in any generation across all problems. We abbreviate chain-of-thought blocks with \texttt{[...]} to highlight the structural elements: reasoning, tool calls, tool responses, and the final Lean proof. The system message is shown in Appendix~\ref{app:tool_prompt}.

\paragraph{Case 1: Wilson's theorem for finite fields.}
The model must prove that the product of nonzero elements of a finite field $K$ is $-1$. It queries \LeanSearch{} for the product of units in a finite field and retrieves \texttt{FiniteField.prod\_univ\_units\_id\_eq\_neg\_one}, which it applies directly.

\begin{Verbatim}[frame=single, fontsize=\scriptsize, framesep=2mm]
<|im_start|>system
[System prompt from Appendix A]<|im_end|>

<|im_start|>user
Complete the following Lean 4 code:

```lean4
import Mathlib
open Function Fintype Subgroup Ideal Polynomial Submodule Zsqrtd
open scoped BigOperators

/-- Let K be a finite field. Prove that the product of the
nonzero elements of K is -1.-/
theorem exercise_13_6_10 {K : Type*} [Field K] [Fintype K*] :
  (Finset.univ.prod fun x : K* => x) = -1 := by sorry
```

Before producing the Lean 4 code to formally prove the given theorem, provide a
detailed proof plan outlining the main proof steps and strategies. The plan should
use the leansearch tool to find relevant theorems from Mathlib that could help
prove this theorem formally.<|im_end|>

<|im_start|>assistant
<think>
I need to prove that the product of the nonzero elements of a finite field K
is -1. This is a generalization of Wilson's Theorem for finite fields.

[... exploration of cyclic group structure and Wilson's theorem ...]

Let me check Mathlib for theorems related to the product of elements in a
finite field.
</think>

<tool_call>
{"name": "leansearch", "arguments": {"query": "product of nonzero elements
of finite field"}}
</tool_call><|im_end|>

<|im_start|>user
<tool_response>
{"tool": "leansearch", "theorems": [
  "FiniteField.prod_univ_units_id_eq_neg_one
      [CommRing K] [IsDomain K] [Fintype K*] :
      prod univ (fun x : K* => x) = (-1 : K*)",
  ...
]}
</tool_response><|im_end|>

<|im_start|>assistant
<think>
I found FiniteField.prod_univ_units_id_eq_neg_one which directly states
the result. Since K is a Field, it is a CommRing and IsDomain; Fintype K*
is given. All typeclass conditions are satisfied.

[... verification of argument ordering and theorem applicability ...]
</think>

```lean4
import Mathlib
open Function Fintype Subgroup Ideal Polynomial Submodule Zsqrtd
open scoped BigOperators

theorem exercise_13_6_10 {K : Type*} [Field K] [Fintype K*] :
  (Finset.univ.prod fun x : K* => x) = -1 := by
  have h_main := FiniteField.prod_univ_units_id_eq_neg_one
  apply h_main
```<|im_end|>
\end{Verbatim}

\paragraph{Case 2: Fermat's theorem on local extrema.}
The model must show that for a smooth non-negative function $f$ with $f(0)=0$ and $f(1)=1$, some iterated derivative is zero. It recognizes that $f(0)=0$ with $f(x) \geq 0$ implies a local minimum at~$0$, queries \LeanSearch{} for the relevant derivative theorem, and retrieves \texttt{IsLocalMin.deriv\_eq\_zero}.

\begin{Verbatim}[frame=single, fontsize=\scriptsize, framesep=2mm]
<|im_start|>system
[System prompt from Appendix A]<|im_end|>

<|im_start|>user
Complete the following Lean 4 code:

```lean4
import Mathlib
open scoped BigOperators

/-- Let f: R -> R be infinitely differentiable with f(0)=0, f(1)=1,
and f(x) >= 0. Show there exist n and x such that f^(n)(x) = 0.-/
theorem exercise_2018_a5 (f : R -> R) (hf : ContDiff R top f)
  (hf0 : f 0 = 0) (hf1 : f 1 = 1) (hf2 : forall x, f x >= 0) :
  exists (n : N) (x : R), iteratedDeriv n f x = 0 := by sorry
```

Before producing the Lean 4 code to formally prove the given theorem, provide a
detailed proof plan outlining the main proof steps and strategies. The plan should
use the leansearch tool to find relevant theorems from Mathlib that could help
prove this theorem formally.<|im_end|>

<|im_start|>assistant
<think>
Since f(0) = 0 and f(x) >= 0 for all x, the point 0 is a local minimum.
By Fermat's theorem on stationary points, the first derivative at a local
minimum must be zero. So taking n=1 and x=0 should work.

[... exploration of why f(0)=0 with global non-negativity implies a local
minimum, and discussion of alternative approaches via Rolle's theorem ...]

Let me search for the relevant Mathlib theorem about derivatives at local
minima.
</think>

<tool_call>
{"name": "leansearch", "arguments": {"query": "derivative is zero at local
minimum"}}
</tool_call><|im_end|>

<|im_start|>user
<tool_response>
{"tool": "leansearch", "theorems": [
  "IsLocalMin.deriv_eq_zero
      (h : IsLocalMin f a) : deriv f a = 0",
  "IsLocalMin.hasDerivAt_eq_zero
      (h : IsLocalMin f a) (hf : HasDerivAt f f' a) : f' = 0",
  "IsLocalMin.fderiv_eq_zero
      (h : IsLocalMin f a) : fderiv R f a = 0",
  ...
]}
</tool_response><|im_end|>

<|im_start|>assistant
<think>
The theorem IsLocalMin.deriv_eq_zero states that the derivative at a local
minimum is zero. I need to show 0 is a local minimum: since f(0) = 0 and
f(x) >= 0 for all x, we have f(x) >= f(0) everywhere. Then
iteratedDeriv 1 f 0 = deriv f 0 = 0 by IsLocalMin.deriv_eq_zero.

[... verification of filter_upwards approach for IsLocalMin ...]
</think>

```lean4
import Mathlib
open scoped BigOperators

theorem exercise_2018_a5 (f : R -> R) (hf : ContDiff R top f)
  (hf0 : f 0 = 0) (hf1 : f 1 = 1) (hf2 : forall x, f x >= 0) :
  exists (n : N) (x : R), iteratedDeriv n f x = 0 := by
  have h_local_min : IsLocalMin f 0 := by
    filter_upwards [isOpen_Iio.mem_nhds (by norm_num : (0 : R) < 1)]
      with x hx
    linarith [hf2 x, hf0]
  have h_deriv_zero : deriv f 0 = 0 :=
    IsLocalMin.deriv_eq_zero h_local_min
  exact <1, 0, by simp [iteratedDeriv_one, h_deriv_zero]>
```<|im_end|>
\end{Verbatim}

\section{Full BFCL per-category results}
\label{app:bfcl_full}

Table~\ref{tab:bfcl_full} provides the complete per-category BFCL v4 results for all 32B models.

\begin{table}[h]
  \centering
  \caption{Full BFCL v4 per-category results for 32B models (accuracy, \%).}
  \label{tab:bfcl_full}
  \scriptsize
  \begin{tabular}{@{}lrrrrrr@{}}
    \toprule
    Category & \Qwen{} & \Goedel{} & +100 & +1k & +18k & +120k \\
    \midrule
    \multicolumn{7}{l}{\textit{Non-live}} \\
    simple\_python       & 95.75 &  10.75 & 93.50 & 92.00 & 91.00 & 83.75 \\
    simple\_java         & 62.00 &   2.00 & 63.00 & 65.00 & 55.00 & 56.00 \\
    simple\_javascript   & 74.00 &   0.00 & 68.00 & 68.00 & 62.00 & 58.00 \\
    multiple             & 96.00 &   9.00 & 92.50 & 94.00 & 88.50 & 87.00 \\
    parallel             & 93.00 &   3.50 & 86.50 & 85.50 & 82.50 & 75.50 \\
    parallel\_multiple   & 91.50 &   1.50 & 81.50 & 79.50 & 80.50 & 80.50 \\
    irrelevance          & 80.00 &  95.83 & 21.67 & 36.25 & 49.17 & 67.08 \\
    \midrule
    \multicolumn{7}{l}{\textit{Live}} \\
    live\_simple         & 87.98 &  10.85 & 83.33 & 83.33 & 82.17 & 77.91 \\
    live\_multiple       & 80.72 &   5.13 & 70.85 & 71.04 & 67.43 & 64.96 \\
    live\_parallel       & 81.25 &   6.25 & 68.75 & 62.50 & 75.00 & 75.00 \\
    live\_par.~mult.     & 66.67 &   0.00 & 54.17 & 50.00 & 66.67 & 58.33 \\
    live\_irrelevance    & 73.42 &  95.14 & 20.25 & 33.48 & 40.16 & 62.67 \\
    live\_relevance      & 93.75 &   6.25 &100.00 & 93.75 & 93.75 & 75.00 \\
    \midrule
    \multicolumn{7}{l}{\textit{Multi-turn}} \\
    base                 & 63.50 &   0.00 & 17.50 & 27.50 &  9.50 &  2.00 \\
    miss\_func           & 50.00 &   0.00 &  4.00 &  5.50 &  0.50 &  0.50 \\
    miss\_param          & 38.50 &   0.00 &  4.00 &  4.50 &  1.50 &  0.50 \\
    long\_context        & 33.50 &   0.00 &  7.00 & 11.50 &  6.50 &  1.50 \\
    \midrule
    \multicolumn{7}{l}{\textit{Memory}} \\
    memory\_kv           & 16.13 &   0.00 &  4.52 &  5.16 &  3.23 &  0.65 \\
    memory\_rec\_sum     & 37.42 &   0.00 &  2.58 & 13.55 &  6.45 &  3.23 \\
    memory\_vector       & 23.87 &   0.00 &  3.23 & 14.84 &  1.29 &  1.94 \\
    \bottomrule
  \end{tabular}
\end{table}

\section{Full BFCL per-category results for the post-RL checkpoint}
\label{app:bfcl_rl_full}

Table~\ref{tab:bfcl_rl_full} provides the complete per-category BFCL v4 results for the post-RL checkpoint \GoedelRL{} and its reactivation variants.
The base model scores near zero on multi-turn and memory categories, consistent with its inability to produce valid tool calls.

\begin{table}[h]
  \centering
  \caption{Full BFCL v4 per-category results for \GoedelRL{} (post-RL checkpoint) and reactivation variants (accuracy, \%).}
  \label{tab:bfcl_rl_full}
  \scriptsize
  \begin{tabular}{@{}lrrrr@{}}
    \toprule
    Category & \GoedelRL{} & +100 & +1k & +18k \\
    \midrule
    \multicolumn{5}{l}{\textit{Non-live}} \\
    simple\_python       & 46.50 & 94.75 & 93.50 & 93.50 \\
    simple\_java         & 29.00 & 65.00 & 62.00 & 57.00 \\
    simple\_javascript   & 16.00 & 72.00 & 66.00 & 64.00 \\
    multiple             & 25.00 & 94.00 & 92.00 & 94.00 \\
    parallel             & 13.50 & 88.00 & 89.50 & 88.50 \\
    parallel\_multiple   &  3.50 & 84.00 & 87.50 & 81.50 \\
    irrelevance          & 91.25 & 21.67 & 33.75 & 34.58 \\
    \midrule
    \multicolumn{5}{l}{\textit{Live}} \\
    live\_simple         & 38.37 & 85.27 & 86.43 & 84.11 \\
    live\_multiple       & 24.03 & 79.01 & 79.87 & 77.11 \\
    live\_parallel       & 12.50 & 62.50 & 75.00 & 75.00 \\
    live\_par.~mult.     &  0.00 & 50.00 & 58.33 & 58.33 \\
    live\_irrelevance    & 82.92 & 21.04 & 29.52 & 33.37 \\
    live\_relevance      &  6.25 &100.00 &100.00 & 93.75 \\
    \midrule
    \multicolumn{5}{l}{\textit{Multi-turn}} \\
    base                 &  0.00 & 19.00 & 25.00 & 14.00 \\
    miss\_func           &  0.00 &  9.50 &  6.50 &  1.50 \\
    miss\_param          &  0.50 &  9.50 &  7.00 &  1.50 \\
    long\_context        &  0.00 & 10.50 & 15.50 &  5.50 \\
    \midrule
    \multicolumn{5}{l}{\textit{Memory}} \\
    memory\_kv           &  0.00 &  0.00 &  1.29 &  0.65 \\
    memory\_rec\_sum     &  0.00 &  1.94 &  7.74 &  1.29 \\
    memory\_vector       &  0.00 &  0.00 &  5.81 &  0.65 \\
    \bottomrule
  \end{tabular}
\end{table}

\end{document}